\documentclass[sigconf, nonacm]{acmart}
\AtBeginDocument{%
  \providecommand\BibTeX{{%
    \normalfont B\kern-0.5em{\scshape i\kern-0.25em b}\kern-0.8em\TeX}}}

\usepackage{microtype}
\usepackage{graphicx}
\usepackage{multicol}
\usepackage{multirow}
\usepackage{cancel}

\newcommand\blfootnote[1]{%
  \begingroup
  \renewcommand\thefootnote{}\footnote{#1}%
  \addtocounter{footnote}{-1}%
  \endgroup
}

\copyrightyear{2024}
\acmYear{2024}
\setcopyright{acmlicensed}\acmConference[WWW '24 Companion]{Companion Proceedings of the ACM Web Conference 2024}{May 13--17, 2024}{Singapore, Singapore}
\acmBooktitle{Companion Proceedings of the ACM Web Conference 2024 (WWW '24 Companion), May 13--17, 2024, Singapore, Singapore}
\acmDOI{10.1145/3589335.3651512}
\acmISBN{979-8-4007-0172-6/24/05}

\begin{document}

\title{Generator-Guided Crowd Reaction Assessment}


\author{Sohom Ghosh}
\affiliation{%
  \institution{Jadavpur University}
  \city{Kolkata}
  \country{India}
  }
\email{sohom1ghosh@gmail.com}

\author{Chung-Chi Chen}
\affiliation{%
  \institution{AIST}
  \city{Tokyo}
  \country{Japan}}
\email{c.c.chen@acm.org}

\author{Sudip Kumar Naskar}
\affiliation{%
  \institution{Jadavpur University}
  \city{Kolkata}
  \country{India}}
\email{sudip.naskar@gmail.com}

\renewcommand{\shortauthors}{Ghosh, et al.}

\begin{abstract}
  In the realm of social media, understanding and predicting post reach is a significant challenge. 
This paper presents a \textbf{C}rowd \textbf{Re}action \textbf{A}ssess\textbf{M}ent (CReAM) task designed to estimate if a given social media post will receive more reaction than another, a particularly essential task for digital marketers and content writers. 
We introduce the Crowd Reaction Estimation Dataset (CRED), consisting of pairs of tweets from The White House with comparative measures of retweet count. 
The proposed Generator-Guided Estimation Approach (GGEA) leverages generative Large Language Models (LLMs), such as ChatGPT, FLAN-UL2, and Claude, to guide classification models for making better predictions. 
Our results reveal that a fine-tuned FLANG-RoBERTa model, utilizing a cross-encoder architecture with tweet content and responses generated by Claude, performs optimally. 
We further 
use a T5-based paraphraser to generate paraphrases of a given post and 
demonstrate GGEA's ability to predict which post will elicit the most reactions. 
We believe this novel application of LLMs provides a significant advancement in predicting social media post reach.
\end{abstract}

\ccsdesc[300]{Information systems~Information retrieval}
\ccsdesc[500]{Computing methodologies~Information extraction}
\ccsdesc[300]{Computing methodologies~Natural language generation}
\ccsdesc[500]{Information systems~Social networks}

\keywords{Large Language Models, Social Media, Natural Language Processing, Crowd Reaction Assessment}



\maketitle

\section{Introduction}
In the swiftly evolving landscape of social media, possessing the capability to accurately anticipate the impact of a post is invaluable. Digital marketers, content creators, and organizations with the intent of influencing or interacting with their audience can strategically enhance their approach and decision-making processes if they can predict post performance prior to its publication. Nevertheless, the challenge at hand, referred to as \textbf{C}rowd \textbf{Re}action \textbf{A}ssess\textbf{M}ent (CReAM), is complicated due to the multifaceted and ceaselessly changing nature of social media engagement.

\blfootnote{© 2024  Copyright held by the owner/author(s). Publication rights licensed to ACM. This is the author's version of the work. It is posted here for your personal use. Not for redistribution. The definitive version was published in WWW ’24 Companion, May 13–17, 2024, Singapore, Singapore, \url{https://doi.org/10.1145/3589335.3651512}}

As social media platforms continue to grow, they become progressively richer data sources reflecting public sentiment and reactions. Despite this, the potential for utilizing such data to predict crowd reactions remains largely uncharted territory. There exists a pressing demand to devise and execute efficacious methods capable of accurately predicting a post's influence, be it in fostering engagement, initiating dialogue, or inspiring action. Comprehending these dynamics not only holds commercial value but also carries significant potential for influencing public discourse and democracy.

\begin{figure}[t] 
    \centering 
    \includegraphics[width=0.475\textwidth]{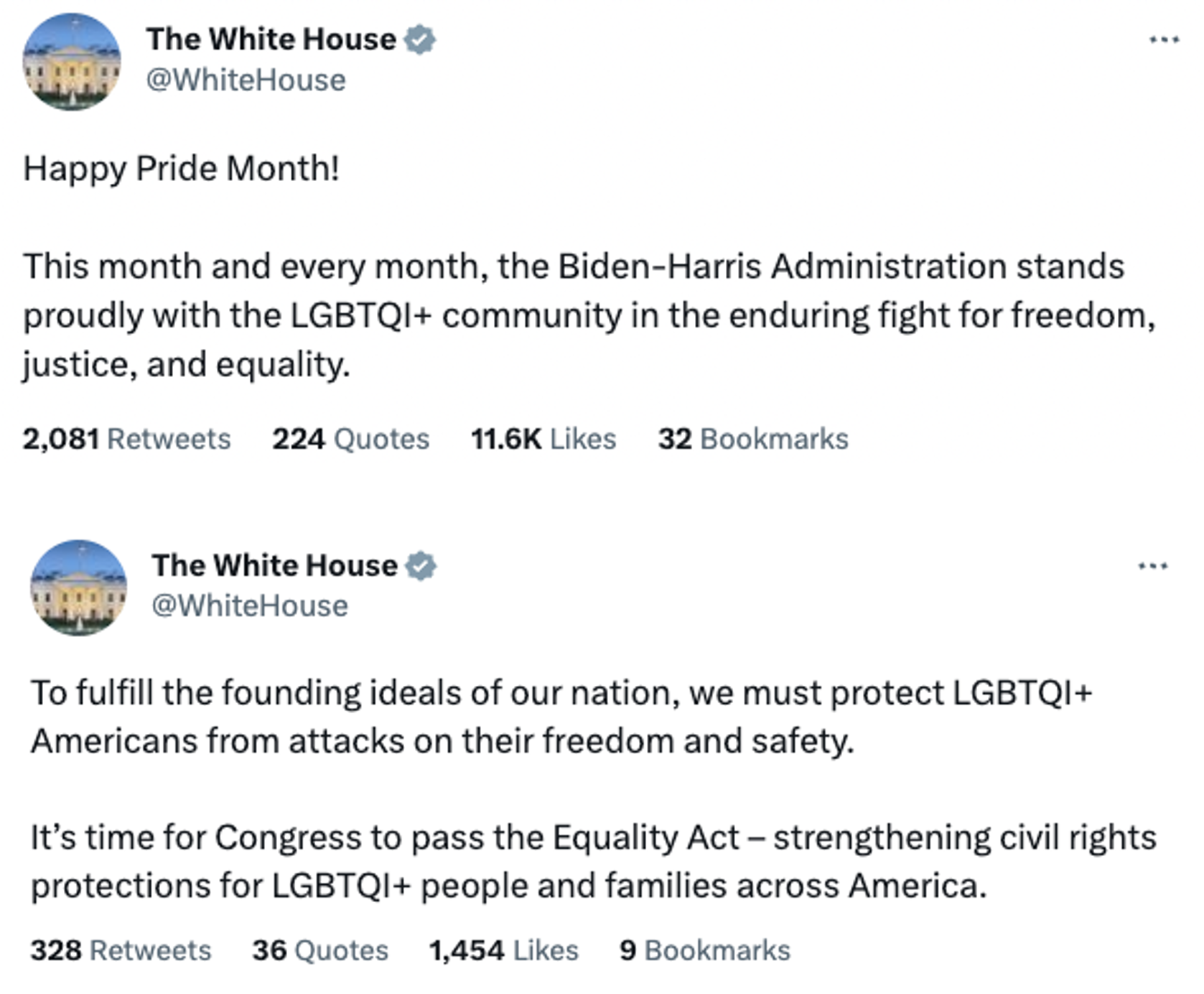} 
    \caption{Example Tweets from The White House.} 
    \label{fig:Example}
\end{figure}

In the process of preparing a social media post, managers often draft multiple versions and select the final one based on an estimation of crowd reactions. As illustrated in Figure~\ref{fig:Example}, when considering two tweets on an identical subject, determining which will receive a greater volume of interactions and responses is a crucial consideration for a social media manager. This paper focuses on tweets issued by The White House and introduces a meticulously curated dataset, the Crowd Reaction Estimation Dataset (CRED).
Specifically, CRED encompasses pairs of tweets paired with comparative metrics of retweet counts.

Recognizing the widespread applications of large language models (LLMs) such as ChatGPT (based on variants of GPT models \cite{gpt3}, FLAN-UL2~\cite{flan2021,tay2023ul2}), and Claude across various sectors
we propose employing their analytical capabilities to address the CReAM task. We suggest a Generator-Guided Estimation Approach (GGEA) that seeks LLMs' analysis on disparate tweets, utilizing the resulting enhanced information to guide classification models in making predictions. The findings from our study indicate that the proposed GGEA contributes to superior performance compared to merely fine-tuning classification models 
and LLMs under one-shot settings. 
The most successful combination involves a fine-tuned FLANG-RoBERTa model \cite{shah-etal-2022-flang} with tweet content and responses generated by Claude.

The major contributions of this paper are as follows:
\begin{itemize}
    \item The introduction of a novel dataset, Crowd Reaction Estimation Dataset (CRED)\footnote{Link for CRED \& codes: \href{https://github.com/sohomghosh/Generator-Guided-Crowd-Reaction-Assessment/}{https://github.com/sohomghosh/Generator-Guided-Crowd-Reaction-Assessment/}}, which simulates the decision-making process of governmental social media managers.

    \item The proposal of a Generator-Guided Estimation Approach (GGEA) that leverages the analytical power of LLMs to predict crowd reactions to social media posts. 
    The demonstration of GGEA's ability to 
    assess paraphrases of a post, providing users with different versions of their content to optimize engagement.
\end{itemize}

\section{Related Work}
The aspiration to maximize audience reach and reactions has been a consistent theme among social media users since the platform's inception. Several studies have focused on understanding and predicting the factors that contribute to the widespread reach of social media posts.
\citet{Cha_Haddadi_Benevenuto_Gummadi_2010} conducted an analysis of the activities of 52 million Twitter users, deducing that a tweet's content is a critical factor in driving retweets.
\citet{suh-2010} assessed 74 million tweets and explored a range of content-based and contextual features that affect the retweet count of a given tweet. They developed a Generalized Linear Model-based model to predict the number of retweets a tweet is expected to generate. Their research underscored the importance of tweet content, URLs, and hashtags as key features in predicting retweet count. Other works on cascade prediction which leverages social networks include \cite{tan-etal-2014-effect} and \cite{deepcas}.

Similarly, \citet{bakshy-2011-everyone} tracked the diffusion activities of 1.6 million Twitter users, 
by studying the relationships between different categories of influencers and the associated cost. Beyond Twitter, research has also been conducted on the reach of Facebook posts. \citet{bernstein-2013-invisible} studied the activities of 222,000 Facebook users, revealing how invisible audiences enhance the reach of Facebook posts. \citet{valkonen2022estimating} investigated how the timing of Facebook posts by Finnish consumers affects reachability.
\citet{tan-etal-2014-effect} analysed the effect of the content of a tweet on its reach. 
More recent studies like \citet{gabriel-etal-2022-misinfo} demonstrated how the reach of news articles can be estimated using models based on GPT-2 \cite{radford2019gpt2} and T5 \cite{raffel2020t5}. However, their approaches provide only a coarse-grained analysis and do not significantly contribute to performing a comparative analysis. They created a dataset by 
annotating reach levels from 1 to 5.

Our work differs from these preceding studies in several significant ways. First, we introduce a new dataset, CRED, that specifically targets the decision-making processes of social media managers in governmental contexts. Second, we propose a novel approach, the GGEA, that employs the analytical power of LLMs to predict crowd reactions to social media posts. Finally, unlike previous works, 
we use a T5-based paraphraser fintuned with ChatGPT responses \cite{chatgpt_paraphraser} over GGEA for
generating and assessing paraphrases of a given post, thereby allowing users to optimize engagement by exploring different versions of their content.

\section{Dataset}

Our primary task is to determine whether a tweet, denoted as $t1$, will receive more retweets than another tweet, denoted as $t2$. Given a pair of tweets, our objective is to predict the tweet with the higher retweet count accurately. 

We compile our dataset from all tweets posted by the official White House Twitter handle from November 2020 to October 2022. We then proceed to curate pairwise data, ensuring each pair adheres to conditions related to temporality, topic, and crowd reaction.

\begin{table}[t]
  \centering
  \small
  \caption{Statistics of CRED. Avg. RT denotes the average number of retweets.}
  \label{tab:Statistics of CRED}%
    \begin{tabular}{lrr}
    \midrule
    \multicolumn{1}{c}{Topic} & \multicolumn{1}{c}{Avg. RT} & \multicolumn{1}{c}{Pairs} \\
    \hline
    Business \& Entrepreneurs & 743.2 & 4,186 \\
    Fitness \& Health & 1010.2 & 628 \\
    Learning \& Educational & 485.1 & 90 \\
    Sports & 728.0 & 6 \\
    \hline
    Total &       & 4,910 \\
    \bottomrule
    \end{tabular}%
\end{table}%

We retained only those instances that fulfill the following four conditions:
\begin{itemize}
    \item The tweets were posted during weekdays: Weekdays generally witness higher levels of social media activity, making them ideal for capturing more interactions and thus providing a more reliable measure of crowd reaction.

    \item The retweet counts of $t1$ and $t2$ differed by at least 10\% of the less re-tweeted tweet. This condition ensures a clear distinction in the crowd reaction between the two tweets, enabling a more accurate and meaningful comparison.

    \item  $t1$ and $t2$ were posted within a gap of 10 days and the difference between time of posting $t1$ and $t2$ was less than or equal to five hours. A short interval between posts helps control the extraneous variables such as time-specific events that might impact retweet counts.

    \item We utilized the Twitter Topic classification model \cite{antypas-etal-2022-twitter} to classify tweets into topics. We considered only those tweet pairs where the model assigned the same topic to both the tweets with a probability of 0.8 or higher.
    
    This high probability threshold ensures that the topics of the tweets are clearly defined and comparable, which is critical for the meaningful analysis of crowd reactions to different tweets on the same topic. 
    
\end{itemize}

Table~\ref{tab:Statistics of CRED} presents the statistics of our proposed Crowd Reaction Estimation Dataset (CRED), and the label distribution in the proposed CRED is well-balanced. 
The average number of retweets across different topics underscores the importance of topic control during dataset construction. 
Topic-wise statistics further highlight that the majority of government tweets pertain to business and entrepreneurship. 
For dataset separation, we use tweets from the initial 1.5 years, i.e., till April 2022, which constitutes 4,304 tweet pairs for training. 
The final 6 months, i.e., May 2022 to October 2022, provide 606 tweet pairs which were used for model validation.

\begin{figure}
\centering
  \includegraphics[width=.45\textwidth]{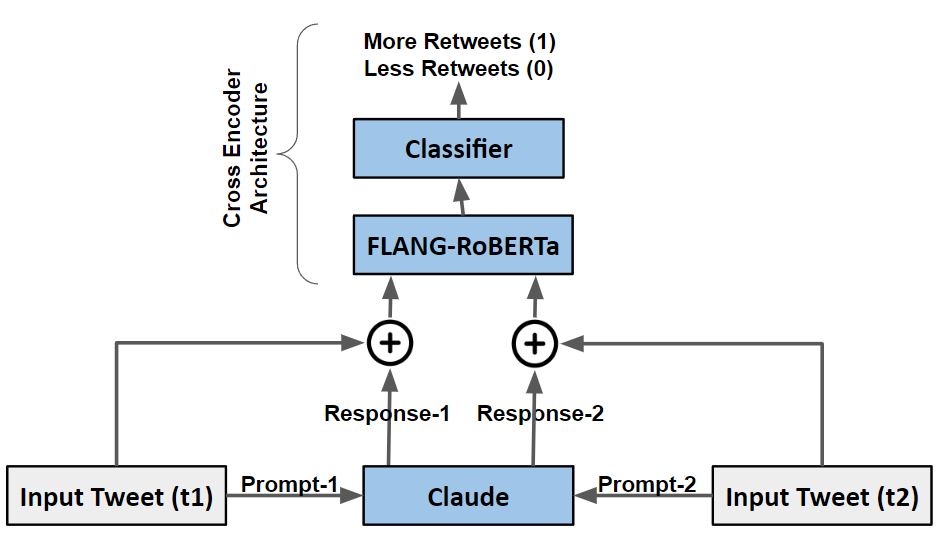}
  \caption{Architecture of GGEA.}
  \label{fig:architecture}
\end{figure}

\section{Methodology}
We present the overall architecture of GGEA in Figure \ref{fig:architecture}. 
The architecture incorporates a FLANG-RoBERTa \cite{shah-etal-2022-flang} model, fine-tuned with the cross-encoder transformer architecture.
The input is a concatenation of the text from a given pair ($t1$,$t2$) and the response obtained from prompting Claude to elucidate why a particular tweet in the pair is more engaging. 
The expected output is whether $t1$ will receive more retweets than $t2$.

To augment usability during inference, we employed an existing open-source T5 \cite{raffel2020t5} based paraphraser \cite{chatgpt_paraphraser} that was trained on paraphrases generated by ChatGPT. 
This paraphraser is utilized to generate alternatives for a given social media post. 
We examine two paraphrases at a time. With the assistance of our developed frameworks, the text predicted to garner maximum reactions is selected from a candidate list comprising the paraphrases and the original post. 
For the Claude part, two types of prompts are employed. 
First, we ask the LLM whether the first tweet will attract more reactions than the second. 
Second, we prompt the LLM with one tweet at a time, querying it to explain the reasons that make the given tweet engaging. 
These responses are either concatenated with the tweets or treated as independent inputs. 

\begin{table}[t]
  \centering
 \caption{Experimental results.}
  \label{tab:Experimental results}%
    \begin{tabular}{llrr}
    \midrule
       \multicolumn{1}{c}{Setup}   & \multicolumn{1}{c}{Model} & \multicolumn{1}{c}{Accuracy} & \multicolumn{1}{c}{F1} \\
    \hline
    Classifier & FLANG-RoBERTa & 50.0\% & 66.7\% \\
    \hline
    \multirow{3}[2]{*}{Zero-Shot} & FLAN-UL2 & 50.2\% & 2.6\% \\
          & Claude & 51.0\% & 9.2\% \\
          & React T5 & 52.3\% & 30.0\% \\
    \hline
    \multirow{3}[2]{*}{GGEA} & GGEA (ChatGPT) & 67.2\% & 68.1\% \\
          & GGEA (FLAN-UL2) & 69.6\% & 70.1\% \\
          & GGEA (Claude) & \textbf{71.9\%} & \textbf{73.1\%} \\
    \hline
    \end{tabular}%
\end{table}%

\section{Experiment} 
\subsection{Experimental Settings}
Our experimental setup compares the outcomes with two categories of baselines: supervised models and zero-shot LLMs. The first heuristic baseline 
fine-tunes FLANG-RoBERTa \cite{shah-etal-2022-flang} to make direct predictions.
To evaluate the feasibility of deploying zero-shot LLMs for the proposed task, we let FLAN-UL2 \cite{flan2021,tay2023ul2}, Claude, and React T5 ~\cite{gabriel-etal-2022-misinfo} make predictions. 
Going one step further, we solicit various LLMs' analysis of $t1$ and $t2$ to determine which LLM can guide the classification model to deliver enhanced performance.
The performance of our models is measured using two metrics: accuracy and F1 score.

\subsection{Experimental Results}
Table~\ref{tab:Experimental results} presents the experimental outcomes. Initial observations reveal that both the classification model and the LLMs struggle to excel in the proposed task. However, it is noteworthy that the proposed GGEA achieves superior performance, irrespective of the LLM consulted. 
We observed that the difference in performance between the GGEA (Claude) i.e. LLM augmented setup and React T5 (the best performing model in zero-shot setup), is significant (p-value$<$0.05). 
This result underlines the importance of collaboration between LLMs and classification models, suggesting that the generated analysis by LLMs successfully augments the information needed for models to make decisions.


Recognizing that different classifiers may influence the performance, we experimented with different classifiers under the GGEA (Claude) setting. Table~\ref{tab:Analysis of changing classifiers in GGEA (Claude)} presents the experimental outcomes. These results suggest that FLANG-RoBERTa outperforms the other tested classifiers.

Finally, topic-wise evaluations are depicted in Table~\ref{tab:Topic-wise evaluation}. We observe that the Learning \& Educational and Sports topics have a limited number of instances. Consequently, we compare the remaining two topics and conclude that GGEA outperforms Fitness \& Health in the Business \& Entrepreneurs topic, primarily due to the larger number of available training instances for the Business \& Entrepreneurs topic.

\begin{table}[t]
  \centering
  \caption{Analysis of changing classifiers in GGEA (Claude).}
  \label{tab:Analysis of changing classifiers in GGEA (Claude)}%
    \begin{tabular}{lrr}
    \toprule
          & \multicolumn{1}{c}{Accuracy} & \multicolumn{1}{c}{F1} \\
    \hline
    GGEA (FLANG-BERT) & 62.7\% & 62.1\% \\
    GGEA (RoBERTa) & 69.8\% & 69.5\% \\
    GGEA (FLANG-RoBERTa) & \textbf{71.9\%} & \textbf{73.1\%} \\
    \bottomrule
    \end{tabular}%
\end{table}%

\begin{table}[t]
  \centering
    \caption{Topic-wise evaluation.  We didn't have any instances for the Sports category in the test set.}
  \label{tab:Topic-wise evaluation}%
    \begin{tabular}{lccc}
    \hline
    \multicolumn{1}{c}{Topic} & \# Instance & Accuracy & F1 \\
    \hline
    Business \& Entrepreneurs & 570 &\multicolumn{1}{r}{73.0\%} & \multicolumn{1}{r}{74.2\%} \\
    Fitness \& Health & 34 & \multicolumn{1}{r}{52.9\%} & \multicolumn{1}{r}{50.0\%} \\
    Learning \& Educational & 2 & \multicolumn{1}{r}{100.0\%} & \multicolumn{1}{r}{100.0\%} \\
    \hline
    \end{tabular}
\end{table}%

\section{Limitations}

This paper demonstrates promising outcomes; however, it is crucial to acknowledge its limitations for a comprehensive understanding of our approach and for guiding future research.

First, the study predominantly utilizes data from a single source, the official Twitter account of the White House. This focus may lead to model overfitting to the unique style, tone, and content characteristic of this account, potentially reducing its applicability to other social media accounts, topics, or platforms. Our preference for retweets over other engagement metrics, such as likes or quote counts, is grounded in the understanding that retweets significantly propagate information \cite{retweet-survey}. The choice of the White House's Twitter account is due to its authenticity, impartiality, and influence on public opinion, along with its wide reach. However, this selection introduces a bias, as engagement levels (e.g., number of retweets) are not directly comparable between accounts with varying follower counts.

Second, numerous factors influence social media engagement, including post timing, social context, influence of preceding posts, visual content, and audience demographics. These aspects are not considered in our current model, which may affect the accuracy of its predictions.

Third, our study is restricted to analyzing retweets as the primary engagement metric on Twitter. Although retweets are a critical engagement indicator, other elements like likes, comments, and shares also contribute to a post's overall impact. Additionally, tweets on different topics are inherently incomparable due to their distinct content. To ensure data quality, we employed a tweet topic classifier \cite{antypas-etal-2022-twitter} with a high-confidence threshold of 0.8, determined empirically. Expanding our model to include these additional engagement metrics would offer a more comprehensive understanding of a post's reach and influence.

\section{Conclusion and Future Directions}
This study provides a novel perspective on predicting crowd reactions to social media posts, introducing a new dataset and a methodology. We curated the Crowd Reaction Estimation Dataset (CRED) which encapsulates the decision-making process of governmental social media managers through the lens of comparative retweet metrics. We also proposed the Generator-Guided Estimation Approach (GGEA), a method that harnesses the analytical prowess of LLMs in a cooperative framework with classification models. Our experiments demonstrate that the GGEA, especially when incorporating Claude's analysis, outperforms mere fine-tuning of classification models and exhibits superior performance in predicting crowd reactions.
These results highlight the potential of LLMs to augment classification models by providing nuanced analysis and further information, thereby enhancing decision-making processes. 

Looking ahead, there are several intriguing avenues for further research. Firstly, GGEA could be expanded to work with other forms of social media, such as Facebook posts or Instagram captions, enabling a more comprehensive understanding of social media engagement across platforms. Secondly, GGEA could be adapted to handle multi-class problems, such as predicting whether a post will go viral or not, rather than just focusing on binary outcomes.
Furthermore, the introduction of more sophisticated paraphrasing methods could potentially enhance the performance of the GGEA. 
Finally, an in-depth exploration of the factors contributing to the performance of different LLMs in the GGEA framework could yield valuable insights. 

\begin{acks}
The work of Chung-Chi Chen was supported in part by JSPS KAKENHI Grant Number 23K16956 and a project JPNP20006, commissioned by the New Energy and Industrial Technology Development Organization (NEDO).
\end{acks}


\balance
\bibliographystyle{ACM-Reference-Format}
\bibliography{sample-base}


\begin{thebibliography}{19}


\ifx \showCODEN    \undefined \def \showCODEN     #1{\unskip}     \fi
\ifx \showDOI      \undefined \def \showDOI       #1{#1}\fi
\ifx \showISBNx    \undefined \def \showISBNx     #1{\unskip}     \fi
\ifx \showISBNxiii \undefined \def \showISBNxiii  #1{\unskip}     \fi
\ifx \showISSN     \undefined \def \showISSN      #1{\unskip}     \fi
\ifx \showLCCN     \undefined \def \showLCCN      #1{\unskip}     \fi
\ifx \shownote     \undefined \def \shownote      #1{#1}          \fi
\ifx \showarticletitle \undefined \def \showarticletitle #1{#1}   \fi
\ifx \showURL      \undefined \def \showURL       {\relax}        \fi
\providecommand\bibfield[2]{#2}
\providecommand\bibinfo[2]{#2}
\providecommand\natexlab[1]{#1}
\providecommand\showeprint[2][]{arXiv:#2}

\bibitem[Antypas et~al\mbox{.}(2022)]%
        {antypas-etal-2022-twitter}
\bibfield{author}{\bibinfo{person}{Dimosthenis Antypas}, \bibinfo{person}{Asahi Ushio}, \bibinfo{person}{Jose Camacho-Collados}, \bibinfo{person}{Vitor Silva}, \bibinfo{person}{Leonardo Neves}, {and} \bibinfo{person}{Francesco Barbieri}.} \bibinfo{year}{2022}\natexlab{}.
\newblock \showarticletitle{{T}witter Topic Classification}. In \bibinfo{booktitle}{\emph{Proceedings of the 29th International Conference on Computational Linguistics}}. \bibinfo{publisher}{International Committee on Computational Linguistics}, \bibinfo{address}{Gyeongju, Korea}.
\newblock


\bibitem[Bakshy et~al\mbox{.}(2011)]%
        {bakshy-2011-everyone}
\bibfield{author}{\bibinfo{person}{Eytan Bakshy}, \bibinfo{person}{Jake~M. Hofman}, \bibinfo{person}{Winter~A. Mason}, {and} \bibinfo{person}{Duncan~J. Watts}.} \bibinfo{year}{2011}\natexlab{}.
\newblock \showarticletitle{Everyone's an Influencer: Quantifying Influence on Twitter}. \bibinfo{publisher}{Association for Computing Machinery}, \bibinfo{address}{New York, NY, USA}.
\newblock
\showISBNx{9781450304931}


\bibitem[Bernstein et~al\mbox{.}(2013)]%
        {bernstein-2013-invisible}
\bibfield{author}{\bibinfo{person}{Michael~S. Bernstein}, \bibinfo{person}{Eytan Bakshy}, \bibinfo{person}{Moira Burke}, {and} \bibinfo{person}{Brian Karrer}.} \bibinfo{year}{2013}\natexlab{}.
\newblock \showarticletitle{Quantifying the Invisible Audience in Social Networks}. \bibinfo{publisher}{Association for Computing Machinery}, \bibinfo{address}{New York, NY, USA}.
\newblock
\showISBNx{9781450318990}


\bibitem[Brown et~al\mbox{.}(2020)]%
        {gpt3}
\bibfield{author}{\bibinfo{person}{Tom Brown}, \bibinfo{person}{Benjamin Mann}, \bibinfo{person}{Nick Ryder}, \bibinfo{person}{Melanie Subbiah}, \bibinfo{person}{Jared~D Kaplan}, \bibinfo{person}{Prafulla Dhariwal}, \bibinfo{person}{Arvind Neelakantan}, \bibinfo{person}{Pranav Shyam}, \bibinfo{person}{Girish Sastry}, \bibinfo{person}{Amanda Askell}, \bibinfo{person}{Sandhini Agarwal}, \bibinfo{person}{Ariel Herbert-Voss}, \bibinfo{person}{Gretchen Krueger}, \bibinfo{person}{Tom Henighan}, \bibinfo{person}{Rewon Child}, \bibinfo{person}{Aditya Ramesh}, \bibinfo{person}{Daniel Ziegler}, \bibinfo{person}{Jeffrey Wu}, \bibinfo{person}{Clemens Winter}, \bibinfo{person}{Chris Hesse}, \bibinfo{person}{Mark Chen}, \bibinfo{person}{Eric Sigler}, \bibinfo{person}{Mateusz Litwin}, \bibinfo{person}{Scott Gray}, \bibinfo{person}{Benjamin Chess}, \bibinfo{person}{Jack Clark}, \bibinfo{person}{Christopher Berner}, \bibinfo{person}{Sam McCandlish}, \bibinfo{person}{Alec Radford}, \bibinfo{person}{Ilya Sutskever}, {and}
  \bibinfo{person}{Dario Amodei}.} \bibinfo{year}{2020}\natexlab{}.
\newblock \showarticletitle{Language Models are Few-Shot Learners}. In \bibinfo{booktitle}{\emph{Advances in Neural Information Processing Systems}}, \bibfield{editor}{\bibinfo{person}{H.~Larochelle}, \bibinfo{person}{M.~Ranzato}, \bibinfo{person}{R.~Hadsell}, \bibinfo{person}{M.F. Balcan}, {and} \bibinfo{person}{H.~Lin}} (Eds.). \bibinfo{publisher}{Curran Associates, Inc.}, \bibinfo{address}{Red Hook, NY, USA}, \bibinfo{pages}{1877--1901}.
\newblock


\bibitem[Cha et~al\mbox{.}(2010)]%
        {Cha_Haddadi_Benevenuto_Gummadi_2010}
\bibfield{author}{\bibinfo{person}{Meeyoung Cha}, \bibinfo{person}{Hamed Haddadi}, \bibinfo{person}{Fabricio Benevenuto}, {and} \bibinfo{person}{Krishna Gummadi}.} \bibinfo{year}{2010}\natexlab{}.
\newblock \showarticletitle{Measuring User Influence in Twitter: The Million Follower Fallacy}.
\newblock \bibinfo{journal}{\emph{Proceedings of the International AAAI Conference on Web and Social Media}} \bibinfo{volume}{4}, \bibinfo{number}{1} (\bibinfo{date}{May} \bibinfo{year}{2010}), \bibinfo{pages}{10--17}.
\newblock
\urldef\tempurl%
\url{https://doi.org/10.1609/icwsm.v4i1.14033}
\showDOI{\tempurl}


\bibitem[Damodaran(2021)]%
        {prithivida2021parrot}
\bibfield{author}{\bibinfo{person}{Prithiviraj Damodaran}.} \bibinfo{year}{2021}\natexlab{}.
\newblock \bibinfo{title}{Parrot: Paraphrase generation for NLU.}
\newblock
\newblock


\bibitem[Firdaus et~al\mbox{.}(2018)]%
        {retweet-survey}
\bibfield{author}{\bibinfo{person}{Syeda~Nadia Firdaus}, \bibinfo{person}{Chen Ding}, {and} \bibinfo{person}{Alireza Sadeghian}.} \bibinfo{year}{2018}\natexlab{}.
\newblock \showarticletitle{Retweet: A popular information diffusion mechanism--A survey paper}.
\newblock \bibinfo{journal}{\emph{Online Social Networks and Media}}  \bibinfo{volume}{6} (\bibinfo{year}{2018}), \bibinfo{pages}{26--40}.
\newblock


\bibitem[Gabriel et~al\mbox{.}(2022)]%
        {gabriel-etal-2022-misinfo}
\bibfield{author}{\bibinfo{person}{Saadia Gabriel}, \bibinfo{person}{Skyler Hallinan}, \bibinfo{person}{Maarten Sap}, \bibinfo{person}{Pemi Nguyen}, \bibinfo{person}{Franziska Roesner}, \bibinfo{person}{Eunsol Choi}, {and} \bibinfo{person}{Yejin Choi}.} \bibinfo{year}{2022}\natexlab{}.
\newblock \showarticletitle{Misinfo Reaction Frames: Reasoning about Readers{'} Reactions to News Headlines}. In \bibinfo{booktitle}{\emph{Proceedings of the 60th Annual Meeting of the Association for Computational Linguistics (Volume 1: Long Papers)}}. \bibinfo{publisher}{Association for Computational Linguistics}, \bibinfo{address}{Dublin, Ireland}.
\newblock


\bibitem[Li et~al\mbox{.}(2017)]%
        {deepcas}
\bibfield{author}{\bibinfo{person}{Cheng Li}, \bibinfo{person}{Jiaqi Ma}, \bibinfo{person}{Xiaoxiao Guo}, {and} \bibinfo{person}{Qiaozhu Mei}.} \bibinfo{year}{2017}\natexlab{}.
\newblock \showarticletitle{DeepCas: An End-to-End Predictor of Information Cascades}. \bibinfo{publisher}{International World Wide Web Conferences Steering Committee}, \bibinfo{address}{Republic and Canton of Geneva, CHE}.
\newblock
\showISBNx{9781450349130}


\bibitem[Radford et~al\mbox{.}(2019)]%
        {radford2019gpt2}
\bibfield{author}{\bibinfo{person}{Alec Radford}, \bibinfo{person}{Jeff Wu}, \bibinfo{person}{Rewon Child}, \bibinfo{person}{David Luan}, \bibinfo{person}{Dario Amodei}, {and} \bibinfo{person}{Ilya Sutskever}.} \bibinfo{year}{2019}\natexlab{}.
\newblock \showarticletitle{Language Models are Unsupervised Multitask Learners}.
\newblock \bibinfo{journal}{\emph{OpenAI blog}} \bibinfo{volume}{1}, \bibinfo{number}{8} (\bibinfo{year}{2019}), \bibinfo{pages}{9}.
\newblock


\bibitem[Raffel et~al\mbox{.}(2020)]%
        {raffel2020t5}
\bibfield{author}{\bibinfo{person}{Colin Raffel}, \bibinfo{person}{Noam Shazeer}, \bibinfo{person}{Adam Roberts}, \bibinfo{person}{Katherine Lee}, \bibinfo{person}{Sharan Narang}, \bibinfo{person}{Michael Matena}, \bibinfo{person}{Yanqi Zhou}, \bibinfo{person}{Wei Li}, {and} \bibinfo{person}{Peter~J. Liu}.} \bibinfo{year}{2020}\natexlab{}.
\newblock \showarticletitle{Exploring the Limits of Transfer Learning with a Unified Text-to-Text Transformer}.
\newblock \bibinfo{journal}{\emph{Journal of Machine Learning Research}} \bibinfo{volume}{21}, \bibinfo{number}{140} (\bibinfo{year}{2020}), \bibinfo{pages}{1--67}.
\newblock


\bibitem[Shah et~al\mbox{.}(2022)]%
        {shah-etal-2022-flang}
\bibfield{author}{\bibinfo{person}{Raj Shah}, \bibinfo{person}{Kunal Chawla}, \bibinfo{person}{Dheeraj Eidnani}, \bibinfo{person}{Agam Shah}, \bibinfo{person}{Wendi Du}, \bibinfo{person}{Sudheer Chava}, \bibinfo{person}{Natraj Raman}, \bibinfo{person}{Charese Smiley}, \bibinfo{person}{Jiaao Chen}, {and} \bibinfo{person}{Diyi Yang}.} \bibinfo{year}{2022}\natexlab{}.
\newblock \showarticletitle{When {FLUE} Meets {FLANG}: Benchmarks and Large Pretrained Language Model for Financial Domain}. In \bibinfo{booktitle}{\emph{Proceedings of the 2022 Conference on Empirical Methods in Natural Language Processing}}. \bibinfo{publisher}{Association for Computational Linguistics}, \bibinfo{address}{Abu Dhabi, United Arab Emirates}.
\newblock


\bibitem[Suh et~al\mbox{.}(2010)]%
        {suh-2010}
\bibfield{author}{\bibinfo{person}{Bongwon Suh}, \bibinfo{person}{Lichan Hong}, \bibinfo{person}{Peter Pirolli}, {and} \bibinfo{person}{Ed~H. Chi}.} \bibinfo{year}{2010}\natexlab{}.
\newblock \showarticletitle{Want to Be Retweeted? Large Scale Analytics on Factors Impacting Retweet in Twitter Network}. \bibinfo{publisher}{IEEE Computer Society}, \bibinfo{address}{USA}.
\newblock
\showISBNx{9780769542119}


\bibitem[Tan et~al\mbox{.}(2014)]%
        {tan-etal-2014-effect}
\bibfield{author}{\bibinfo{person}{Chenhao Tan}, \bibinfo{person}{Lillian Lee}, {and} \bibinfo{person}{Bo Pang}.} \bibinfo{year}{2014}\natexlab{}.
\newblock \showarticletitle{The effect of wording on message propagation: Topic- and author-controlled natural experiments on {T}witter}. In \bibinfo{booktitle}{\emph{Proceedings of the 52nd Annual Meeting of the Association for Computational Linguistics (Volume 1: Long Papers)}}. \bibinfo{publisher}{Association for Computational Linguistics}, \bibinfo{address}{Baltimore, Maryland}.
\newblock


\bibitem[Tay et~al\mbox{.}(2023)]%
        {tay2023ul2}
\bibfield{author}{\bibinfo{person}{Yi Tay}, \bibinfo{person}{Mostafa Dehghani}, \bibinfo{person}{Vinh~Q. Tran}, \bibinfo{person}{Xavier Garcia}, \bibinfo{person}{Jason Wei}, \bibinfo{person}{Xuezhi Wang}, \bibinfo{person}{Hyung~Won Chung}, \bibinfo{person}{Dara Bahri}, \bibinfo{person}{Tal Schuster}, \bibinfo{person}{Steven Zheng}, \bibinfo{person}{Denny Zhou}, \bibinfo{person}{Neil Houlsby}, {and} \bibinfo{person}{Donald Metzler}.} \bibinfo{year}{2023}\natexlab{}.
\newblock \showarticletitle{{UL}2: Unifying Language Learning Paradigms}. In \bibinfo{booktitle}{\emph{The Eleventh International Conference on Learning Representations}}.
\newblock


\bibitem[Valkonen et~al\mbox{.}(2022)]%
        {valkonen2022estimating}
\bibfield{author}{\bibinfo{person}{Lauri Valkonen}, \bibinfo{person}{Jouni Helske}, {and} \bibinfo{person}{Juha Karvanen}.} \bibinfo{year}{2022}\natexlab{}.
\newblock \showarticletitle{Estimating the causal effect of timing on the reach of social media posts}.
\newblock \bibinfo{journal}{\emph{Statistical Methods \& Applications}} (\bibinfo{year}{2022}), \bibinfo{pages}{1--15}.
\newblock


\bibitem[Vladimir~Vorobev(2023)]%
        {chatgpt_paraphraser}
\bibfield{author}{\bibinfo{person}{Maxim~Kuznetsov Vladimir~Vorobev}.} \bibinfo{year}{2023}\natexlab{}.
\newblock \bibinfo{title}{A paraphrasing model based on ChatGPT paraphrases}.
\newblock
\newblock


\bibitem[Wei et~al\mbox{.}(2022)]%
        {flan2021}
\bibfield{author}{\bibinfo{person}{Jason Wei}, \bibinfo{person}{Maarten Bosma}, \bibinfo{person}{Vincent Zhao}, \bibinfo{person}{Kelvin Guu}, \bibinfo{person}{Adams~Wei Yu}, \bibinfo{person}{Brian Lester}, \bibinfo{person}{Nan Du}, \bibinfo{person}{Andrew~M. Dai}, {and} \bibinfo{person}{Quoc~V Le}.} \bibinfo{year}{2022}\natexlab{}.
\newblock \showarticletitle{Finetuned Language Models are Zero-Shot Learners}. In \bibinfo{booktitle}{\emph{International Conference on Learning Representations}}.
\newblock


\bibitem[Xu et~al\mbox{.}(2020)]%
        {xu-etal-2020-autoqa}
\bibfield{author}{\bibinfo{person}{Silei Xu}, \bibinfo{person}{Sina Semnani}, \bibinfo{person}{Giovanni Campagna}, {and} \bibinfo{person}{Monica Lam}.} \bibinfo{year}{2020}\natexlab{}.
\newblock \showarticletitle{{A}uto{QA}: From Databases to {QA} Semantic Parsers with Only Synthetic Training Data}. In \bibinfo{booktitle}{\emph{Proceedings of the 2020 Conference on Empirical Methods in Natural Language Processing (EMNLP)}}. \bibinfo{publisher}{Association for Computational Linguistics}, \bibinfo{address}{Online}.
\newblock


\end{thebibliography}


\newpage

\section*{Appendix}
\appendix

\section{Hyperparameters}
Hyperparameters of all the models we trained are mentioned in Table \ref{tab:hyperparameters}.

\section{Prompts}
\label{sec:Prompts}
Prompt engineering is an emerging field of research. We experimented with different prompts to understand what works for us.
The exact prompts are as follows:\\
\textbf{Type 1: Prompt for FLAN-UL2. Chat-GPT, and Claude:} \\
text-1: <\textbf{\texttt{t1}}> \\
text-2: <\textbf{\texttt{t2}}> \\
Will text-1 receive more reactions than text-2. Answer me with "yes", "no", just one word. \\

\noindent \textbf{Type 2: Prompt for FLAN-UL2} \\
Why is the following text so engaging? \\
Text: <tweet>\\
The text is engaging because\\

\noindent \textbf{Type 2: Prompt for  Chat-GPT, and Claude}\\
Why is the following text so engaging? \\
Text: <tweet>\\

Following guardrails, Chat-GPT did not respond to prompts of Type 1.
Details about the LLMs which were used are mentioned in Table \ref{tab:llm-details}. We observed that for Type 2 prompts, the responses from FLAN-UL2 were always shorter and crisp compared to Claude and ChatGPT. The responses from the FLAN-UL2 model consisted of maximum 24 words and 9 words in average. The Claude model provided us with point-wise answers in most cases, and it often quoted parts from the tweet while explains why the tweet was engaging. The responses from the Claude model (297 words) were generally about twice as long as that of the ChatGPT model (128 words).

\begin{table}[t]
\small                             

\caption{Hyperparameters of the models trained. Wherever nothing is mentioned, we use the default parameters.}
\label{tab:hyperparameters}
\centering
\begin{tabular}{l|l}
\hline
\multicolumn{1}{c}{\textbf{Model Type}} & \multicolumn{1}{c}{\textbf{Hyperparameters}}         \\ 
\hline
RoBERTa & \multirow{3}{*}{\begin{tabular}[c]{@{}l@{}}Epochs=15, Train batch size =8, \\ weight\_decay=0.01,  \\ learning\_rate=2e-5\end{tabular}} \\ 
FLANG -RoBERTa      &                                  \\ 
FLANG-BERT          &                                  \\ \hline
Cross Encoder       & Epochs=20, Train batch size = 16 \\ 
\hline
\end{tabular}
\end{table}

\begin{table*}[]
\caption{Details about the LLMs used.}
\label{tab:llm-details} 
\centering
\begin{tabular}{l|l}
\hline
\multicolumn{1}{c}{\textbf{LLM}} & \multicolumn{1}{c}{\textbf{Details}}                                              \\ \hline
FLAN-UL2     & Model: 20B parameters, \url{https://huggingface.co/google/flan-ul2} \\ \hline
Claude & \begin{tabular}[c]{@{}l@{}}Model: claude-v1, max. tokens to sample=200,\\ \url{https://www.anthropic.com/index/introducing-claude}\end{tabular} \\ \hline
ChatGPT       &  \begin{tabular}[c]{@{}l@{}}Model: gpt-3.5-turbo [Based on variants of GPT models \cite{gpt3}], \\ \url{https://openai.com/blog/chatgpt}\end{tabular}       \\ 
\hline
\end{tabular}
\end{table*}

\section{Paraphrasers}
We explored three most downloaded open source paraphrasers present in the Hugging Face, which had descriptions of the underlying model architectures. We qualitatively assessed open-source paraphrasers based on their ability to generate complete and sensible paraphrases. However, we did not conduct any quantitative assessment. More details about the paraphrasers are presented in Table~\ref{tab:paraphrases}. We finally selected the T5-based paraphraser \cite{chatgpt_paraphraser} trained using responses from ChatGPT. We use the following hyperparameters: num\_beams=5,
max\_length=128,
num\_return\_sequences=5,
temperature=0.7,
    num\_beam\_groups=5,
    repetition\_penalty=10.0,
    diversity\_penalty=3.0, and
    no\_repeat\_ngram\_size=2.

\begin{table*}[]
\centering
\caption{Output of different paraphrasers for the following input text. \textbf{Input:} \textit{It’s time to rebuild an American economy that works for all of our families and the next generation. It’s time to ensure every American enjoys an equal chance to get ahead. It’s time to build our economy back better.}}
\label{tab:paraphrases}
\begin{tabular}{l|l}
\toprule
\multicolumn{1}{c}{\textbf{Paraphraser}} &
  \multicolumn{1}{c}{\textbf{Output}} \\ \hline
\begin{tabular}[c]{@{}l@{}}prithivida/parrot\_paraphraser\_on\_T5 \\ \cite{prithivida2021parrot}\end{tabular} &
  \begin{tabular}[c]{@{}l@{}}It’s time to rebuild an American economy that \\works for all of our families and the next\end{tabular} \\ \hline
\begin{tabular}[c]{@{}l@{}}stanford-oval/paraphraser-bart-large \\ \cite{xu-etal-2020-autoqa}\end{tabular} &
  \begin{tabular}[c]{@{}l@{}}It is time to rebuild our economy to work for all our \\ families and for the next generation\end{tabular} \\ \hline
\begin{tabular}[c]{@{}l@{}}humarin/chatgpt\_paraphraser\_on\_T5\_base \\ \cite{chatgpt_paraphraser} \end{tabular} &
  \begin{tabular}[c]{@{}l@{}}The time has come to rebuild an American economy \\ that benefits all Americans, including our families and \\ the next generation. It's time to ensure every American \\ has an equal chance to succeed. We need to improve \\ our economy back on track.\end{tabular} \\ \bottomrule 
\end{tabular}
\end{table*}

\section{Other Experiments}
We experimented with several model architectures like RoBERTa-base (Accuracy: 0.55, F1: 0.68), fine-tuned FLANG-RoBERTa embedding using SBERT architecture then passing the embeddings through a feedforward neural network (Accuracy: 0.69, F1: 0.67), etc. However, none of these experiments improved the performance. We experimented with FLANG-RoBERTa as the ``Business \& Entrepreneurs'' category dominates our dataset. FLANG-RoBERTa has been specifically fine-tuned for the finance domain. FLANG-RoBERTa outperformed RoBERTa in the ``Business \& Entrepreneurship'' category. In the ``Fitness \& Health'' category, RoBERTa out-performed FLANG-RoBERTa. Lastly, in the ``Learning \& Educational'' category, performance of both the models is the same.

We collected a sample of 88 instances (pairs of tweets) with a similar number of re-tweets count (within 10\%). We call it Bucket-0. For Bucket-0 the Accuracy is 56.82\% and the F1 score is 56.82\%.
On the test set, for different buckets constructed using the difference in number of retweets (diff), the proposed model performs as follows:\\ 
\textbf{Bucket-1} (10\%<=diff<60\%), Accuracy: 61.2\%, F1: 62.9\%\\
\textbf{Bucket-2} (60\%<=diff<141.3\%), Accuracy: 68.4\%, F1: 68.8\%\\
\textbf{Bucket-3} (141.3\%<=diff<311.5\%),  Accuracy: 80.67\%, F1: 81.52\%\\
\textbf{Bucket-4} (diff>=311.5\%), Accuracy: 77.6\%, F1: 79.0\%

Lastly, we used Large Language Models (like ChatGPT, Claude) as they are well known for their reasoning skills and Pre-trained Languages Models (like BERT, RoBERTa) as they are the state of the art for discriminative tasks. Our motivation was to understand if complementing the discriminative models with inputs from large language models would help. This is like leveraging the best of both worlds. The response from LLMs improved the quality of the features. Thus, the  GGEA architecture outperformed others.

\section{Example}
\begin{figure}[t] 
    \centering 
    \includegraphics[width=0.48\textwidth]{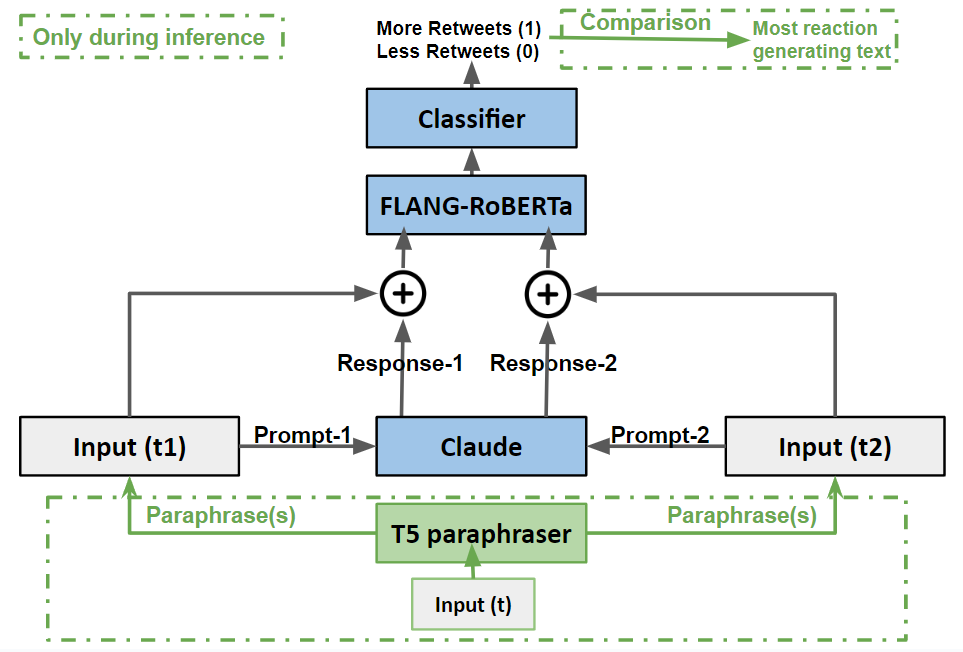} 
    \caption{GGEA along with paraphraser during scoring/inference.} 
    \label{fig:ggea-full}
\end{figure}

In Figure \ref{fig:ggea-full}, we present how the GGEA framework along with the paraphraser during scoring look like.
Here is an example of how GGEA framework works. 

\textbf{Input:} \textit{This year, our economy is projected to grow at the fastest pace in nearly 40 years. Right now, we have the opportunity to make once-in-a-generation investments in the foundations of middle class prosperity. Read more about the American Jobs Plan:} \\

\noindent \textbf{Paraphrases generated}:\\
\textbf{1)} \textit{We are expecting the fastest growth in our economy for almost four decades this year. At present, we have the chance to invest in bolstering middle-class prosperity. Learn more about the American Jobs Plan.} \\
\textbf{2)} \textit{The economy is expected to expand at a rate faster than it has been for almost 40 years this year. We have the opportunity to invest in the middle class's success once again. Learn more about the American Jobs Plan.} \\
\textbf{3)} \textit{We are expecting the fastest growth in our economy for almost four decades this year. At present, we have the chance to invest in bolstering middle-class prosperity. Learn more about the American Jobs Plan: Why it matters?
} \\
\textbf{4)} \textit{The economy is expected to expand at a rate faster than it has been for almost 40 years this year. We have the opportunity to invest in the middle class's success once again. Learn more about the American Jobs Plan.} \\
\textbf{5)} \textit{Our economy is poised to grow at its fastest rate in nearly 40 years this year, making it a prime opportunity for us all to invest in the middle class's success. Learn more about the American Jobs Plan:
} \\

\noindent \textbf{Output:}\\
The paraphrase which is supposed to get maximum retweets as per GGEA is: \\ \textit{The economy is expected to expand at a rate faster than it has been for almost 40 years this year. We have the opportunity to invest in the middle class's success once again. Learn more about the American Jobs Plan.}.

\section{Data and Code Availability}
The datasets along with the codes can be downloaded from Github.\footnote{\url{https://github.com/sohomghosh/Generator-Guided-Crowd-Reaction-Assessment/}} We shall release them under CC BY-NC-SA 4.0 licence after the acceptance of our manuscript.
\end{document}